\def\year{2018}\relax
\documentclass[letterpaper]{article} 
\usepackage{aaai18}  
\usepackage{times}  
\usepackage{helvet}  
\usepackage{courier}  
\usepackage{url}  
\usepackage{graphicx}  

\usepackage{amsmath}
\usepackage{amsfont}

\begin{document}
%
\title{Hi, how can I help you?: Automating enterprise IT support help desks}
\author{Senthil Mani$^1$, Neelamadhav Gantayat$^1$, Rahul Aralikatte$^1$, Monika Gupta$^1$ \\
	\tt \{sentmani, neelamadhav, rahul.a.r, monikgup\}@in.ibm.com \\
	\AND Sampath Dechu$^1$, Anush Sankaran$^1$, Shreya Khare$^1$ \\
	\tt \{sampath.dechu, anussank, shkahre4\}@in.ibm.com \\
	\AND
	Barry Mitchell$^2$, Hemamalini Subramanian$^2$, Hema Venkatarangan$^2$ \\
	\tt bcm@us.ibm.com, \{hesubram, hema.sudarshan\}@in.ibm.com \\
	$^1$IBM Reseach \qquad $^2$IBM Global Business Services
}
\maketitle
\begin{abstract}
Question answering is one of the primary challenges of natural language understanding. In realizing such a system, providing complex long answers to questions is a challenging task as opposed to factoid answering as the former needs context disambiguation. The different methods explored in the literature can be broadly classified into three categories namely: 1) classification based, 2) knowledge graph based and 3) retrieval based. Individually, none of them address the need of an enterprise wide assistance system for an IT support and maintenance domain. In this domain the variance of answers is large ranging from factoid to structured operating procedures; the knowledge is present across heterogeneous data sources like application specific documentation, ticket management systems and any single technique for a general purpose assistance is unable to scale for such a landscape. 
To address this, we have built a cognitive platform with capabilities adopted for this domain. Further, we have built a general purpose question answering system leveraging the platform that can be instantiated for multiple products, technologies in the support domain. The system uses a novel hybrid answering model that orchestrates across a deep learning classifier, a knowledge graph based context disambiguation module and a sophisticated bag-of-words search system. This orchestration performs context switching for a provided question and also does a smooth hand-off of the question to a human expert if none of the automated techniques can provide a confident answer. This system has been deployed across 675 internal enterprise IT support and maintenance projects. 
\end{abstract}

\graphicspath{{./}}

\section{Introduction}
One of the long-term goals of AI is to have a system answering complex questions asked in natural language. The term question answering (QA) has a very broad interpretation enabling almost any problem to be formulated as one. The problem of QA originated decades ago with research focused on retrieving answers from structured databases. AI has aided the growth of QA systems in more challenging settings with better natural data understanding. QA systems have undergone a revolution in inferring from various information sources such as unstructured web content, large-scale document corpus and provide more cognitive responses~\cite{fang2016community}. 

Based on the response provided, QA systems can be broadly classified into two groups: i) factoid based systems, which generally answer `wh' questions. The answer is typically an entity with one or few words~\cite{xu2015longest}~\cite{oh2016semi}, ii) non-factoid based systems providing complex long answers or a procedural explanation to the question. 

Research in factoid QA systems is very mature~\cite{kolomiyets2011survey} with lots of open domain systems available such as AskHermes~\cite{cao2011askhermes}, AnswerBus~\cite{zheng2002answerbus}, QANUS~\cite{ng2015qanus}, YodaQA~\cite{baudivs2015modeling}, Sirius~\cite{hauswald2015sirius}, and DEQA~\cite{lehmann2012deqa}. Publicly available large-scale knowledge bases, such as \textit{FreeBase}, are manually curated to learn and benchmark factoid based QA systems~\cite{yao2014information}~\cite{bordes2015large}.  

However, going beyond factoid QA systems, i.e. retrieving long answers is extremely challenging both in defining what constitutes an answer chunk and in evaluating the retrieval performance~\cite{yang2016beyond}. In such non-factoid systems, generating syntactically correct long procedural answers is challenging, compelling the use of classification or search based approaches for such tasks~\cite{clark2016combining}~\cite{mitra2015addressing}. 

In enterprise IT, one of the main sources of income is providing support to clients. In 2012, the IT support industry revenue in the U.S. alone was estimated to be \$212 billion~\cite{shapiro2014us}. This is an area where even simple automations can bring about a huge financial impact.~\cite{chen2012business} studied how intelligent systems can make this sector more profitable. In this work, we present a platform which is developed and deployed inside IBM to share the load of IT support agents.

This system is a classical case of a QA system where the answers are not factoid but are long explanatory answers which often contain multiple steps describing the solution to a problem. The success of such a system in enterprise depends on many factors like: i) accuracy of the answers ii) speed of answer generation/retrieval iii) ability to rapidly ingest new data.

Consider this example from the banking domain: ``\textit{How can I get a card?}". It is the responsibility of a system to respond incomplete/generic questions. To do this, we need a robust technique which will just not try to answer the question right away but asks for more information to build a clear context. Any single technique will not be able to accomplish this and therefore an ensemble of different techniques are required to work together to understand the question, identify the gaps in the information given by the user and then answer the question in a meaningful manner. To cater to such situations, we have built a platform which can do the following with no assistance from a human.

\begin{enumerate}
	\item \textbf{Knowledge ingestion:} The system has the ability to assimilate knowledge from multiple sources. The sources currently supported are unstructured documents (\textit{.docx, .pptx, .pdf, .html}), images, audio and video recordings.
	\item \textbf{Dynamic Disambiguation:} If the input question is incomplete and/or extra information is required for answering the question with high confidence,  the system asks a counter question to the user rather than provide a generic answer
	\item \textbf{Orchestration:} There are multiple components in the system which can provide the answer in different situations. Therefore, we have built a separate orchestrator which decides which component to call under what circumstances to maximize the likelihood of getting a correct answer
	\item \textbf{Effective addition of human-in-the-loop}: Even with all our efforts, there will always be questions that a machine can never answer. To handle such cases, the orchestration mechanism is built with the ability to automatically hand-off a question to a human agent who can then carry forward the conversation started by the system
	\item \textbf{Continuous learning:} The system never stops learning. It monitors i) the user's feedback after every dialog turn and ii) the conversations between the user and the system/agent to improve its knowledge store
\end{enumerate}

Further, once the answer has been identified, if there is a back-end fix which needs to be performed to completely resolve the problem, the platform automatically executes an appropriate automaton with the user's permission. For example, if the user's problem was that he is not able to log into a system, the platform, after narrowing down to a high confidence solution, asks the consumer whether it should reset his password. On receiving an affirmative response, the reset process is performed, hence automating the end-to-end IT technical support process. This platform is currently being used by close to $1300$ internal enterprise IT support and maintenance projects.

\section{System architecture}

\begin{figure*}[t]
	\center
	\includegraphics[scale=.5]{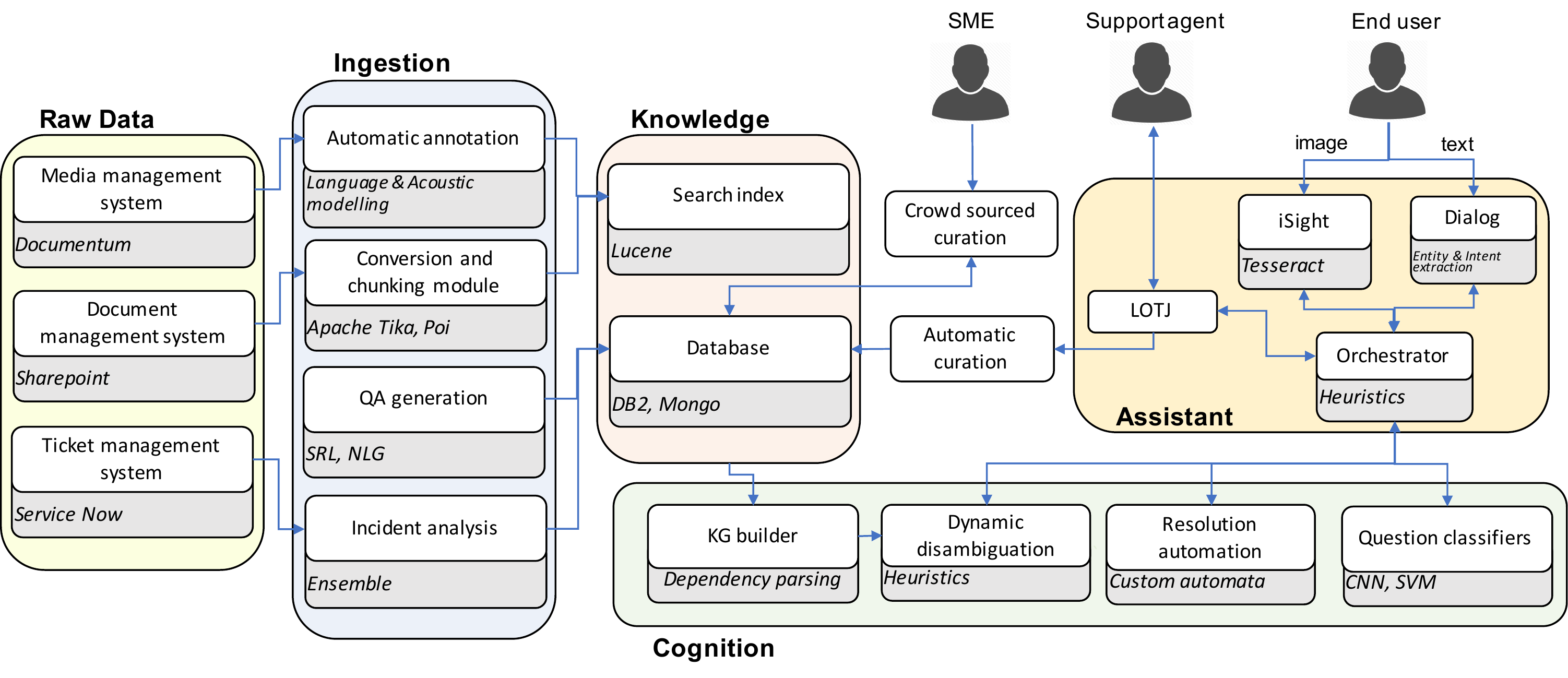}
	\caption{\label{arch}Overall architecture diagram of our platform}
\end{figure*}

This platform consists of three major components: (i) an \textbf{ingestion} component which converts \textbf{raw data} in various formats into structured \textbf{knowledge}, (ii) \textbf{cognitive} component which learns and indexes the answer for a question through multiple techniques, and (iii) the \textbf{assistant} component which interacts with the end user and orchestrates across the cognitive component to fetch the answer. In this section, we discuss these components in detail

\subsection{Knowledge Curation}
Enterprise knowledge is dispersed across multiple system (incident management systems, document management systems etc.) and formats (html, pdf, ppt, doc etc.).  In some cases, the knowledge resides only with the SMEs. To build a question answering system, ingestion of this knowledge is fundamental. The knowledge curation component addresses this heterogeneity by enabling a crowd-sourced platform for curating QA pairs manually, mining incident logs to extract QA pairs, and understanding and extracting content from various structured and unstructured documents.   

\subsubsection{Incident Analysis}
The success of a question answering system largely depends on understanding what are the frequently asked questions by the users and being able to address these questions. Past incidents in incident management systems can be mined to understand this insight. However, incidents contain this data (the problem reported and resolution provided) in an unstructured format. Further the quality of this data is human dependant and also prone to human error. 

We implemented a clustering approach to mine frequently occurring problem categories by grouping individual incidents to sets of clusters. We then extracted the most frequent problem in the cluster as the question and the corresponding resolution (if available) as the relevant answer. However, in some cases, we had to rely on manual curation of the content to create a relevant QA pair from an identified problem cluster. 

\subsubsection{Crowd Sourced Curation}
Much of the knowledge in resolving commonly occurring user issues, is present with the SMEs who have been working in this domain. To bootstrap the system quickly with clean, noise-free QA pairs, we enabled a crowd sourced QA authoring component in our platform. SMEs could use this component to author the answers for the commonly occurring user issues. 

\subsubsection{Document Ingestion}
The purpose of document ingestion is to break content into atomic chunks such that each chunk explains about one concept. The document ingestion pipeline handles \textit{.doc}, \textit{.docx}, \textit{.ppt}, \textit{.pptx}, and \textit{.pdf} files in two major steps. In the first step, Apache POI\footnote{https://poi.apache.org/} is used to check if the document has good formatting i.e. the content is structured with headings and sub-headings. If they are not formatted in this way, we use some structural heuristics and Latent Semantic Indexing~\cite{LSI} to induce formatting into the documents. The structured document in then converted into HTML using Apache TIKA\footnote{https://tika.apache.org/}. 
Post processing is performed on the HTML chunks to fix any discrepancies or incorrect chunking. The processed chunks are then indexed using Apache Lucene\footnote{http://lucene.apache.org/} to enable fast retrieval. This pipeline maintains the styles, formatting, tables, images, etc in the documents so that they can be displayed in an aesthetically pleasing manner to the user.

\subsubsection{Multi-Media Ingestor}
In an enterprise, there exist typically knowledge transfer sessions about applications in recorded video formats. These are sessions conducted by an outgoing vendor to the service provider receiving the maintenance contract of the client IT systems. However, knowledge contained in such videos remain inaccessible because it is laborious for the support agent to get the relevant information, as these videos stretch over couple of hours of knowledge transfer sessions. Further, these videos generally show a screen capture of a running application or programs and the related audio contains the relevant correlated information. 

Our ingestor component is an ensemble of audio processing techniques which enables support agent to retrieve relevant video snippets based on a query. The main aim of this component is to convert the information contained in the video to a richly annotated text format so that it can be indexed in the same way as documents. The implementation details are listed below. 


\begin{enumerate}
\item \textbf{Noise Removal:} Adaptive filtering~\cite{sambur1978adaptive} is used to remove the noise present in the audio channel which may have seeped in due to the recording environment, microphone processing algorithms, etc.

\item \textbf{Speech To Text:} 
The filtered audio is then converted into time stamped text using Automatic Speech Recognition (ASR). With the application of deep learning techniques, ASR performance has improved dramatically with word error rate dropping to about $6\%$ from $14\%$ a few years ago~\cite{deng2013new}. We use our proprietary speech to text service which has been shown to have a state-of-the-art word-error rate of $5.5\%$~\cite{speech_totext}. We use WaveNet~\cite{oord2016wavenet} and a custom LSTM for language modelling, and an ensemble of three neural networks for acoustic modelling viz., i) an LSTM with multiple feature inputs, ii) an LSTM trained with speaker-adversarial multi-task learning and iii) a residual net (ResNet) with $25$ convolutional layers and time-dilated convolutions. This process produces word-level time stamped text \textit{without punctuations} from the audio content.

\item \textbf{Transcript Enrichment:} The transcripts obtained from the previous step lack punctuation, and sentence level time-stamps. Lack of punctuations limits the use of these transcripts for subsequent machine processing like machine translation, question answering etc. It is impossible to perform phrase level (contiguous and non-contiguous words) or topic based searches if we cannot identify or segment the obtained transcript into sentences. Therefore, we trained a bi-directional recurrent neural network (BRNN) model with attention similar to~\cite{tilk2016bidirectional}. It takes in a sequence of words with no punctuation and outputs punctuated text. 

\item \textbf{Indexing and Search} Finally, the transcripts are indexed with Lucene. This allows us to indirectly search the content of videos using its transcripts. We use Lucene's highlights and Extended Dismax Query Parser (edismax) to retrieve contiguous and non-contiguous text (and hence video) fragments respectively. This allows the user to have a seamless experience of playing the video from the exact moment where a topic of interest starts. 


\end{enumerate}

\subsubsection{Question \& Answer Generation}

\begin{figure}[t]
	\center
	\includegraphics[scale=0.5]{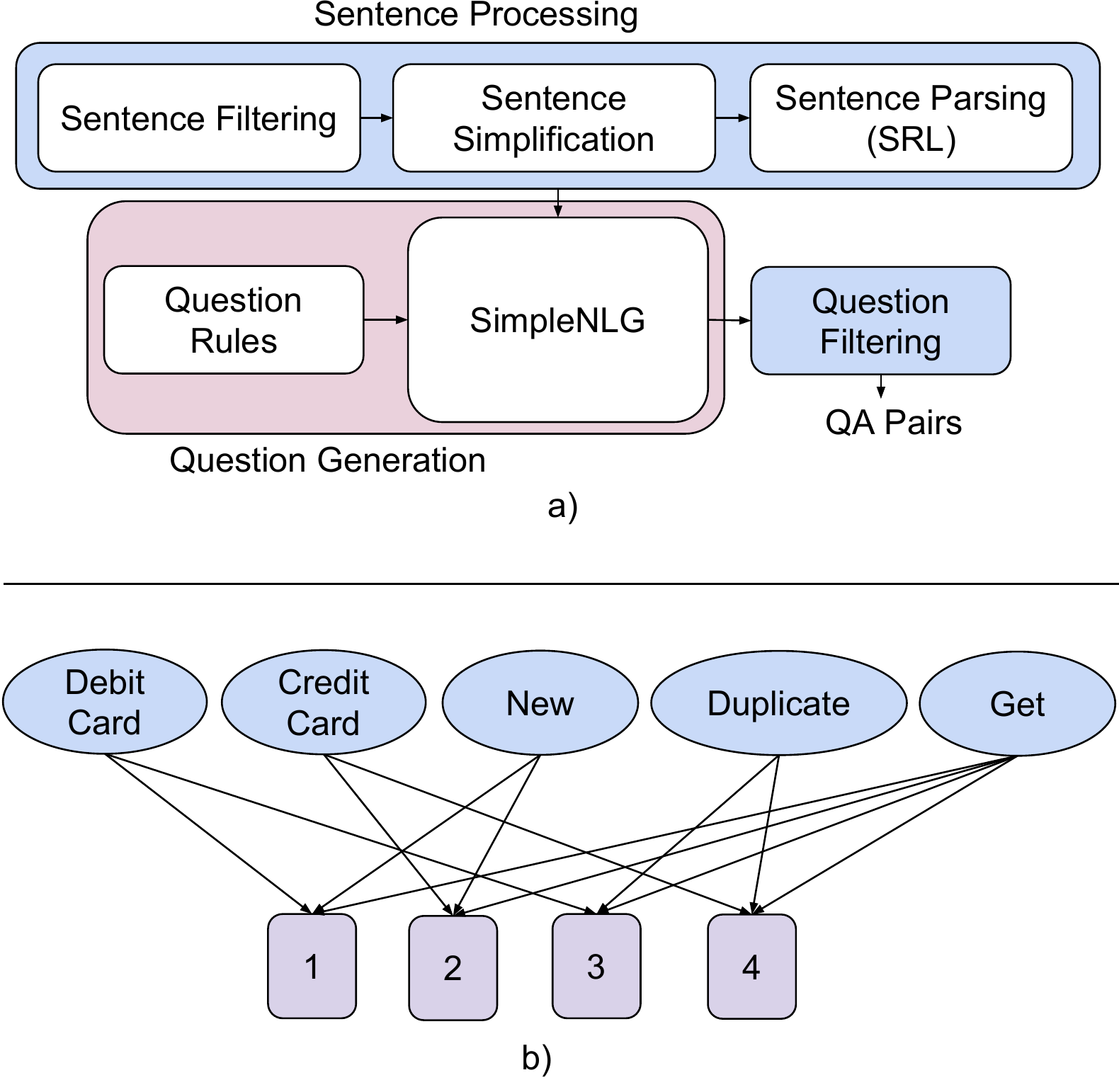}
	\caption{\label{fig:qa_kg} a) Question generation pipeline. b) Knowledge graph created on the resultant questions.}
\end{figure}

One of the important goals of our system is to reduce the effort of manual curation. Hence, it is essential to identify QA pairs directly from the documents and videos that are ingested for a project. Figure~\ref{fig:qa_kg}a shows the various sub-components of a QA pair generation pipeline. Currently, we create questions from `factual' sentences and `mention' sentences. \textit{Factual sentences} are those which contain one or more atomic facts. For eg: ``A credit card is a standalone lending instrument.''. \textit{Mention sentences} are those which talk about a location such as ``the following table...'', ``the above figure...'', etc. We briefly explain each subcomponent of this pipeline below.

\begin{enumerate}
\item\textbf{Sentence filtering:}
We identify important sentences from a chunk of text which are candidates for QA pair generation. We achieve this by first creating an extractive summary using TextRank algorithm~\cite{mihalcea2004textrank} retaining $10\%$ of the sentences. Further, we filter the sentence using a domain specific glossary constructed automatically by using tf-idf scores from the domain corpus. 

\item\textbf{Sentence simplification:}
The candidate sentences can sometime be complex in nature. Such sentences are broken into simpler ones using ~\cite{heilman2010extracting}, such that the truth value of the simpler sentences are always true. 

 
\item\textbf{Sentence parsing:}
The candidate set of sentences are parsed using Semantic Role Labeling (SRL)~\cite{pradhan2004shallow} to get the constituents. For each predicate in a sentence, further identified constituents which are either a semantic role (agent, patient, instrument, etc.) or an adjunct (location, manner, temporal etc.).  We implemented the SRL using 
a SVM~\cite{cortes1995support} trained on the WSJ corpora from the PTB dataset~\cite{marcus1994penn}. These semantic labels are used to form grammatically correct questions following the rules described next.

\item\textbf{Question formation rules:} 
We identify the question type (Who, What, How, When etc.) to be generated for each constituent of a predicate based on the combination of semantic role and the named entity class . For e.g if the word is a constituent, the role is an agent and named entity class is a person, then we generate a \textit{Who} question. We constructed $42$ such static rules (similar to ~\cite{mannem2010question}) based on the combination of constituent, role and named entity class.  



\item\textbf{Question Generation:} 
For each candidate sentence, we use SimpleNLG\footnote{\url{https://github.com/simplenlg}} an open-source library for generating natural language question based on thier SRL annotations and the type of question identified in the previous steps.


\item\textbf{Question Filtering:} 
The questions generated can sometimes be grammatically incorrect or too generic and not relevant to the domain. We use some heuristics such as length of the question, presence of a noun phrase, etc. to retain high quality questions and discard others. 

\end{enumerate}

\subsection{Cognitive Component}

In this section, we talk about the core cognitive component of the platform. This layer uses the knowledge ingested in the previous stages to answer questions from the users. There are three components in this layer, each of which takes a different approach to tackle the user's questions.
\begin{itemize}
	\item \textbf{QA classifiers}: We can look at QA as a question classification problem. One or more questions in the ingested QA pairs can have the same answer. The first question for which a new answer is written/ingested becomes a `primary question` and all other questions which have the same answer become it's `alternate questions'. Each primary question and it's corresponding alternate questions form a class. The purpose of these classifiers is to classify incoming questions into one of these classes. A simple SVM and an advanced CNN classifier are built to suit the needs of different users
	\item \textbf {Dynamic Disambiguation:} Sometimes the users' questions are not complete. Additional information/context is required before the classifiers can classify the question correctly with a high confidence score. A sophisticated disambiguation module is built to deal with such cases of incomplete data. A knowledge graph is built using the ingested data and is used by the disambiguation module to identify the gaps in the user's question and ask for relevant information
	\item \textbf{Search}: If a question cannot be classified with confidence even after getting missing information from the user, a simple tf-idf and bag of words based language model is used to search for similar sentences and retrieve them
\end{itemize}

\subsubsection{QA Classifiers}
There are two types of classifiers in our platform which can be configured for a project based on users' specific needs: i) Dependency based CNN classifier (DCNN) and ii) Support Vector Classifier (SVC). There is a trade off between time taken to train and accuracy between these two classifiers. If there is need at the project level, to have an incoming question/answer pair to be available immediately as part of the classifier (especially when the new answer is introducing a new class), then SVC is preferred as it can be trained in a couple of minutes. However, if the project is comfortable with a delay in the incoming question/answer pair's availability in the classifier, then DCNN is preferred. The details of the classifiers are explained below.


\begin{enumerate}
\item\textbf{Dependency based CNN Classifier}
We used CNNs because~\cite{2016arXiv160906686R} showed that CNNs give better classification accuracy on short pieces of text when compared to RNNs. The DCNN classifier is similar to the model proposed in~\cite{DBLP:journals/corr/Kim14f}, but it also considers the ancestor and sibling words in a sentence's dependency tree. Here, we consider two types of convolutions: \textit{Convolutions of ancestor paths} and \textit{Convolutions on Siblings}. For a detailed explanation of these convolutions, please see~\cite{DBLP:journals/corr/MaHZX15}.

We use max-over-time pooling ~\cite{DBLP:journals/corr/Kim14f} to get the maximum activation over the feature maps. In DCNNs, we want the maximum activation from the feature map across the whole dependency tree (whole sentence). This is also called `max-over-tree' pooling~\cite{DBLP:journals/corr/MaHZX15}. Thus, each filter outputs only one feature vector after pooling. The final representation of a sentence will be many such features, one from each of the filters in the network. These feature vectors are finally passed on to a fully connected layer for classification.

The model we built is similar to what is done in~\cite{DBLP:journals/corr/MaHZX15}. We concatenated the feature maps obtained from ancestor path and sibling convolutions with the sequential n-gram representation. The concatenation with the sequential n-gram representation was required since the questions that the model is being trained on may contain grammatical flaws which will result in parsing errors during dependency tree construction, but these parsing errors do not affect the sequential representation.

\item\textbf{Support Vector Classifier}
A standard Support Vector Classifier with a linear kernel is used with a one-vs-rest scheme for multi-class support. We trained it with L2 penalty and hinge loss. We built it using the $liblinear$ library~\cite{Chang:2011:LLS:1961189.1961199} written in C. This makes it extremely fast to train and deploy.

\end{enumerate}

\subsubsection{Dynamic Disambiguation}
Consider an example from the banking support domain, where a customer asks a question like `How do I get a card?'. This is a very generic question and the QA classifier classifies it into multiple semantically related classes such as: 
 \begin{enumerate}
 	\item How can I get a new debit card?
 	\item What is the process to get a new credit card?
 	\item I want a duplicate debit card
 	\item How do I get a duplicate credit card?
 \end{enumerate} 
The confidence scores of these classifications are not high as all these questions are similar to each other. Therefore, when the highest confidence from the classifiers go below a configurable threshold, the dynamic disambiguation module is invoked. The user is asked to clarify the intent with a question like `Do you need a Debit Card or Credit Card?'. Based on the response, further disambiguating question like `Do you want to apply for a new card or a duplicate card?' is posted. Once these relevant information are obtained, the context becomes clear and the classifier can easily identify the relevant QA pair and display the correct answer.
 
To perform such disambiguations, we constructed a knowledge graph (KG) on the fly using the top-k QA pair returned by the classifiers. The KG is a bipartite graph where all the concepts like noun phrases, verb phrases, etc. form one group and the references to the questions form another group as shown in~\ref{fig:qa_kg}b. These concepts are extracted using dependency parsing~\cite{de2008stanford}~\cite{de2008stanford2}. Similar concept buckets are identified from the first group and one question is asked for each bucket. The questions are usually very simple and hence are template based. Once the user clarifies the concept(s) she is interested in, the answer which is connected to those concept(s) is shown.


\subsubsection{Search}
A tf-idf based search system is built on top of Apache Lucene. Lucene offers an $O(n)$ search time, where the text chunks (from the document and AV ingestion pipelines) are indexed using a simple bag-of-words scheme. For an input question, the ten most similar documents/videos along with the closely related chunks are extracted by checking for maximum cosine similarities on their tf-idf representations.

\subsubsection{Resolution Automation}
Sometimes it is not enough to just provide a solution to a user's problem. For eg., if the user is facing an issue during login, it would be a great user experience if the password is reset by the system directly, rather than giving out the instructions for the user to do the same. Therefore our platform supports an automaton to be attached to an answer. When displaying such an answer, the user is alerted to the possibility of the system taking some corrective action on their behalf. If the user agrees, the automaton is executed by the system and the problem is resolved.

\subsection{Assistant}
The main purpose of the Assistance layer is to hide the complexities of the system from the users and provide them with an easy to use interface. The user can interact with our platform in two ways: text and images. The user can ask a question or can upload a screenshot of the error he is facing. The output of the system is always rich text. This layer contains five components, namely:
\begin{enumerate}
	\item \textbf{iSight}: If the user chooses to upload an error screenshot, this component extracts information about the error and certain UI artifacts which may help in resolving the error
	\item \textbf{Dialog}: This is a standard dialog module which takes in user's natural language inputs and converts it into a structured and annotated format understandable by the orchestrator. It also takes in output from the orchestrator and generates natural language text for the user to read
	\item \textbf{Orchestractor}: As the name suggests, this component orchestrates among all the components of the brain to get a single answer and show it to the user
	\item \textbf{LOTJ}: This \textbf{L}earning \textbf{O}n \textbf{T}he \textbf{J}ob component constantly listens to the interactions between the user, the orchestrator and the support agent to identify new tidbits of information which is ingested into the system as new knowledge usable in future conversations
\end{enumerate}

\begin{figure}[t]
	\center
	\includegraphics[scale=.25]{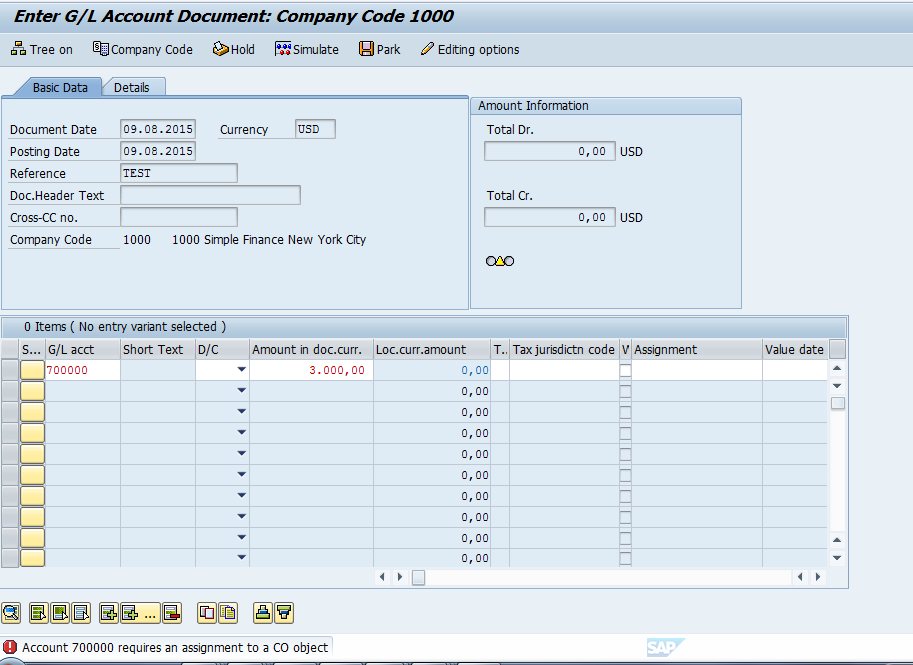}
	\caption{\label{template1}An example screenshot image of a SAP system having an error message displayed. The user query input to find a solution for the error was, ``\textit{How to fix error in CO object?}".}
\end{figure}

\begin{figure}[t]
	\center
	\includegraphics[scale=.35]{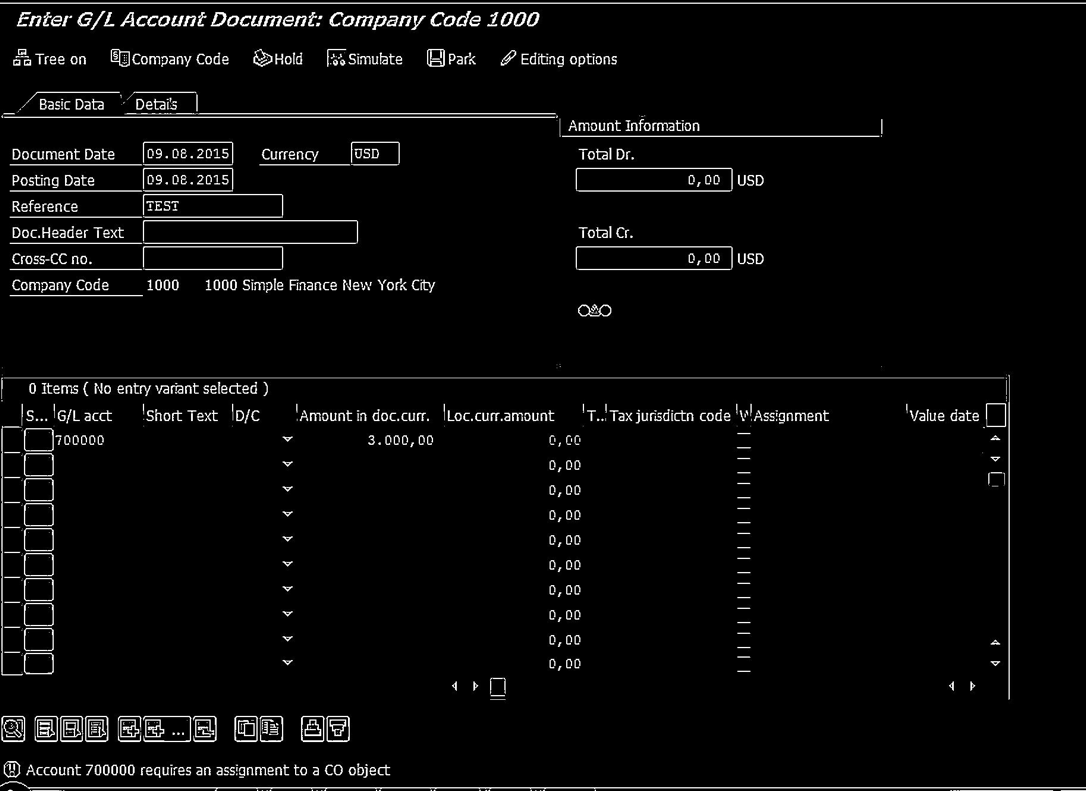}
	\caption{\label{template2}The resulting image of the proposed image preprocessing and enhancement pipeline on which the Tesseract OCR is performed.}
\end{figure}

\subsubsection{iSight}
Sometimes it is easier for the users to upload the screenshot of an error they are facing rather than describing it in text. Consider Figure~\ref{template1} which shows an error in an SAP system UI (the red fields). It may be difficult for a novice user to describe this error and hence might end up asking a generic question like ``\textit{How to fix error in CO object?}". This will result in a unnecessary disambiguation process in this context, which can be avoided if the system accepts the error screen shot directly and parses them to identify the errors accurately.

We used the Tesseract OCR~\cite{smith2007overview} library to extract text from an image. Upon initial testing, we found out that the performance of Tesseract's text extraction is influenced by, (i) font size, (ii) font color, (iii) background color, and (iv) font type. Hence, we built the following image preprocessing pipeline to enhance the text in screenshot images: i) Converted RGB color to grayscale for withering the effect of font and background colors ii) Removed the effect of font types such as bold and italics, by sharpening the text using two levels of Difference of Gaussians iii) Normalized the font size by enhancing the resolution of the images (super-resolution) using bicubic interpolation iv) converted the image into a binary format by performing adaptive thresholding using Otsu's algorithm and v) completed the binarization loss through one level of dialation using a square structuring element.

The resulting enhanced screenshot image is shown in Figure~\ref{template2} from which text is extracted using Tesseract. A general purpose English dictionary along with a domain specific vocabulary built using ingested documents is used to correct spellings in the OCR extracted text. Further, a simple classifier is used to recognize the application to which this UI screenshot belongs to and heuristics are applied to extract the error message and meta information. This is then restructured appropriately and fed into the orchestrator for further processing.

\subsubsection{Dialog}
This component does a two way conversion between natural language and a structured data format understandable by the system. It extracts the intent in the user's question and the relevant entities the question is talking about. These intents are entities are then processed and enriched using a fuzy matching algorithm~\cite{baeza1999faster} before being sent to the orchestrator.

\subsubsection{Orchestrator}
The orchestrator component acts as a middleman between the user and the platform. It communicates with all the components in the cognitive component to get relevant answers and decides what to show to the user. It also gathers feedback from the user at every step to identify it's own shortcomings. The major steps taken by the orchestrator when a new conversation starts with a user is as follows: i) The user either uploads a screenshot or enters a question which is then pre-processed to a structured format and given to the orchestrator ii) The orchestrator initially contacts the classifiers to get top-k answers for the question asked and their confidence scores iii) If the confidence of an answer is greater than a threshold, the answer is shown. Else, the dynamic disambiguation module is called and the orchestrator asks counter questions to the user to clarify the context iv) After clarification, if the confidence of the classifiers increase above the threshold, the answer is shown to the user. Else, the search module is invoked vi) If the high confidence answer shown to the user has a automaton attached to it, the orchestrator takes the user's permission and executes the automaton vii) At any time, if the user is not happy with the answer provided, the orchestrator does a smooth hand-off to a human agent

\subsubsection{Learning on the job}
The purpose of this module is to identify new QA pairs which can be harvested from interactions with the system. It does so in two ways: i) whenever the search module is invoked and the user provides a positive feedback on a document/AV chunk, the LOTJ module associates the question asked by the user to this chunk and forms a new QA pair and updates the associated weight based on the frequency. If this weight becomes greater than a pre-defined threshold, the QA pair is ingested into the system ii) when the user is dissatisfied with the answers given by the system and a hand-off is made to the human agent, the LOTJ component listens to the conversation between the user and the agent to identify potential QA pairs, updates the associated weights and the same process is followed as before.

At any point in time, the support agents can see the potential QA pairs created by the LOTJ module. They can then edit them if required and approve/reject them for immediate ingestion. Thus, this module provides a non-intrusive way of accumulating new knowledge.

\section{Business Impact }
Our platform has enabled the deployment of multiple general purpose assistance solutions to reduce the time taken to resolve incidents, automate the resolution of incidents and reduce incident volumes resulting in support agent productivity improvement. Some of the solutions built on our platform are: 
 \begin{enumerate}
 	\item Employee Assist: Assistance to business users; helping them tackle day-to-day issues they face while interacting with various applications (like HR portal, Travel portal, etc) and there by reducing the overall volume of incidents raised
 	\item Agent Assist: Assistance to support agents (L1/L2) in helping them resolve incidents faster
 	\item Coding Assist: Assistance to SAP ABAP programmers to help improve their productivity in development of custom ABAP modules
 	\item Dynamic Automation: Automating the execution of typical service requests orchestrating across robotic process automatons, thereby reducing the need for a human in the loop.
 \end{enumerate} 
 
These solutions have been deployed across multiple accounts and helped significantly in revenue generation and savings. Table~\ref{summary} summarizes current usage of these solutions in a production environment.  
\begin{table}[t]
	\begin{center}
		\begin{tabular}{|l | l | }
			\hline
			\bf Metric  & \bf Value \\
			\hline \hline
			Projects  & 675     \\   \hline 
			Users  & 5800     \\ \hline
			Avg. users / project  & 86     \\ \hline
			Total QA Pairs  & 306930     \\ \hline
			Avg. QA pairs / project  &   454.7   \\ \hline
			Queries per day  & 453     \\ \hline
			Avg. time saved / user  & 455 Minutes     \\ \hline
			\multicolumn{2}{c}{\textbf{Questions answered by}} \\ \hline
			a) Classifiers  & 106033     \\ \hline
			b) Dynamic Disambiguation  & 12590     \\ \hline
			c) Search  & 113351     \\
			\hline
		\end{tabular}
	\end{center}
	\caption{\label{summary} Usage summary of the solutions in production}
\end{table}

Here are some actual quotes from the user community who have actively adopted our solutions: 

 \begin{enumerate}
 	\item \textit{``Using Agent Assist we are able to resolve tickets faster. This one implementation has given a lot of mindshare with the client and is a key ingredient in our contract renewal proposal with the client''}
	\item \textit{``Over 30+ support agents were able to use the solution to resolve 30\% of the tickets. Saving time on ticket resolution means helping end users quicker, and reducing the effort and cost required to support the application in question. Win-win!'}
	\item \textit{``We are thrilled to have a successful commercial deployment of the automation solution. This kind of automation is becoming a fundamental way GBS delivers services, a definite stepping stone towards self healing applications and wider process integration''}
 \end{enumerate} 

We have a dedicated development team of around 30 people including testers as part our services business units who manage this product following an agile development methodology. There are monthly releases consisting of new enhancements and bug fixes, following our own devops pipeline to build, test and deploy the solutions to production. 

\section{Conclusion}
We have developed a cognitive QA platform aimed at automating enterprise help desks for IT support and maintenance. A generic orchestration framework is designed that integrates diverse QA methods and performs context switching in real-time to combine the advantages of the individual methods. We also discuss knowledge ingestion, training \& configuration, the user experience and the continuous learning \& adaptation aspects of this system. We also emphasize on how including a human expert in the loop guarantees a high-quality user experience. Currently, the QA system provides support to $675$ enterprise projects, thereby improving support agent productivity while reducing the turnaround time for incident resolution and also attracting a huge revenue.

\bibliography{ref}
\bibliographystyle{aaai}

\end{document}